\documentclass[11pt]{article}

\usepackage[]{acl}
\usepackage{times}
\usepackage{latexsym}
\usepackage[T1]{fontenc}
\usepackage[utf8]{inputenc}
\usepackage{microtype}
\usepackage{inconsolata}
\usepackage{graphicx}
\usepackage{subcaption}
\usepackage{booktabs}
\usepackage{amsmath}
\usepackage{amssymb}
\usepackage{mathtools}
\usepackage{amsthm}
\usepackage{multirow}
\usepackage{makecell}
\usepackage{float}

\usepackage[capitalize,noabbrev]{cleveref}

\theoremstyle{plain}

\theoremstyle{definition}

\theoremstyle{remark}

\usepackage{bm}

\DeclareMathOperator*{\argmax}{arg\,max}

\newcommand{\mbzuailogo}{\raisebox{-0.2\height}{\includegraphics[height=1.0em,trim=15.12bp 15.48bp 260.04bp 15.24bp,clip]{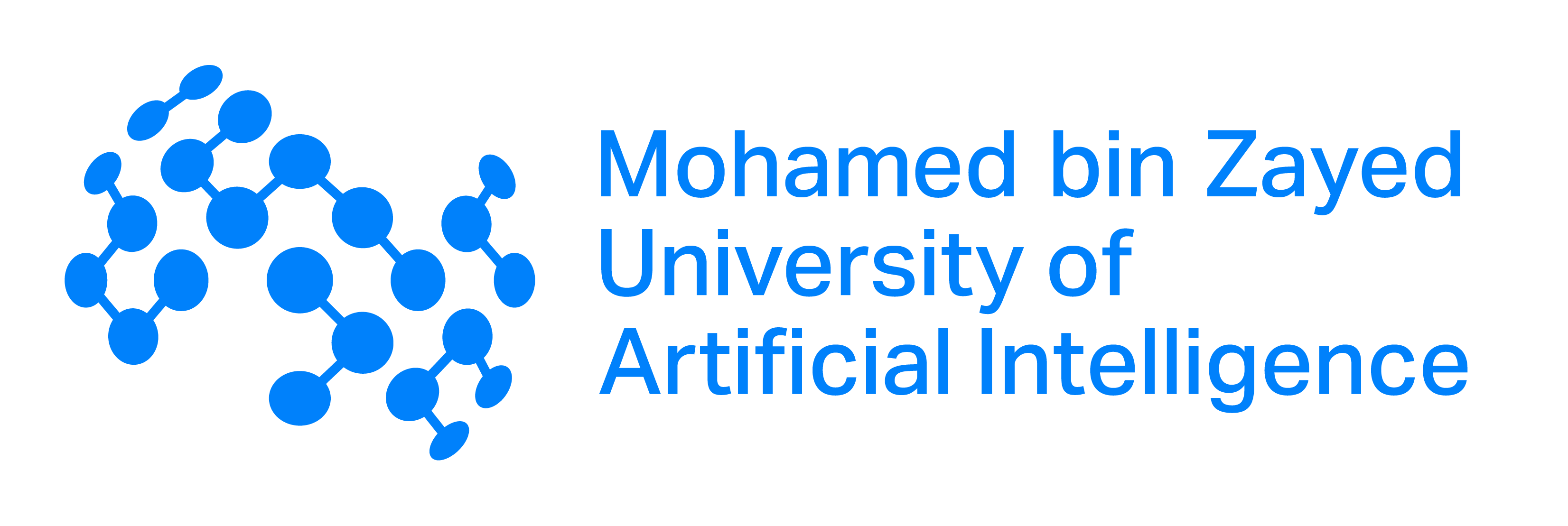}}\,}

\title{Sparse-to-Sparse Feature Learning for Cross-Layer Analysis and Steering}
\author{Masahiro Kaneko \quad Ayana Niwa \quad Timothy Baldwin \\
  \mbzuailogo MBZUAI \\
  \texttt{\{Masahiro.Kaneko, Ayana.Niwa, Timothy.Baldwin\}@mbzuai.ac.ae}}

\begin{document}
\maketitle

\begin{abstract}

Understanding how models internally represent information is a central challenge in interpretation and control.
Sparse autoencoders (SAEs) decompose dense hidden representations into sparse features, and cross-layer methods capture structure spanning multiple layers.
However, existing objectives do not explicitly train the sparse code at one layer to predict the sparse code at the next layer.
We propose sparse-to-sparse feature learning (\textbf{S2S}), where sparse features at each intermediate layer are trained to predict those of the immediately following layer, with the final layer anchored by an output-distribution loss and optimization proceeding sequentially from output to input. 
Experiments across eight language models show stronger adjacent-layer correspondence than matched sparse-feature baselines. We evaluate feature-based interventions at matched output-distribution shift, separating steering-induced effects from sparse-pathway reconstruction effects. Causal ablation provides complementary evidence beyond correspondence metrics.
  
\end{abstract}

\section{Introduction}
Large language models (LLMs) have achieved remarkable capabilities across diverse tasks~\citep{brown2020language, grattafiori2024llama, anthropic2025claude4, comanici2025gemini, hurst2024gpt}.
Despite these successes, understanding how these models internally represent information remains a critical challenge for safety and controllability~\citep{bereska_mechanistic_2024}.
A key obstacle to interpretability is the dense nature of hidden representations in neural networks: each neuron participates in encoding multiple features, and each feature is distributed across many neurons, a phenomenon known as superposition~\citep{elhage2022toy}.
This dense and entangled encoding makes it difficult to isolate and interpret individual features encoded within intermediate representations.

Sparse autoencoders (SAEs) have emerged as a promising approach to address this challenge.
By learning more features than input dimensions under sparsity constraints, SAEs decompose dense hidden states into interpretable sparse feature vectors.
However, since SAEs are primarily trained to reconstruct within a single layer, tracking their contributions to subsequent layers remains difficult.
Transcoders~\citep{dunefsky2024transcoders}, cross-layer transcoders (CLTs)~\citep{ameisen2025circuit}, and Crosscoders~\citep{lindsey2024crosscoders} explicitly capture cross-layer structure, and e2e-SAE~\citep{braun2024e2e} additionally optimizes features for output behavior.
These methods are important precursors to our work. Our narrower distinction is at the training target: they reconstruct or predict dense hidden states, combine information across layers, or optimize vocabulary outputs, whereas S2S explicitly predicts an adjacent layer's sparse code.

This distinction is critical for interpretability. 
When correspondences between sparse features across layers are not explicitly modeled, it becomes difficult to trace how individual features are transformed or preserved as information propagates through the network.
A natural alternative would be to impose hard constraints, such as using a shared sparse autoencoder across layers~\citep{shi2025routesae}. 
However, such constraints may limit expressivity and fail to capture the layer-specific abstractions known to emerge in LLMs.

In this work, we propose \textit{sparse-to-sparse} feature learning (\textbf{S2S}), a method that learns the relationships between sparse features across layers.
In S2S, sparse feature vectors at each intermediate layer are trained to predict the sparse feature vectors of the immediately following layer. The final layer is trained against the model's output distribution, and optimization proceeds sequentially from the output layer toward the input layer. 
By learning relationships directly in sparse space, S2S imposes a soft constraint that encourages alignment between corresponding features across layers while allowing each layer to maintain its own representational structure. We use cross-layer analysis to mean following features through chains of such adjacent-layer correspondences.

We evaluate S2S on eight LLMs along three complementary dimensions. 
First, to directly assess whether S2S aligns sparse features across layers, we measure the consistency of feature activation patterns between adjacent layers.
Second, we evaluate feature-based interventions using held-out external task metrics at matched output-distribution shift.
Third, we test output preservation and causal feature ablation.

Our experiments show that S2S improves adjacent-layer consistency, preserves output-relevant information, and concentrates causal importance in high-activation features. We evaluate intervention efficiency separately at matched output-distribution shift.
Because correspondence metrics partially overlap with the training objective, we treat the intervention and ablation results as the main independent evidence. We do not equate structural correspondence alone with human interpretability.

\section{Sparse-to-Sparse Feature Learning}
\label{sec:our_method}

We propose \textbf{S2S}, a method that learns sparse features by propagating output-relevant information backward through successive sparse feature spaces.
Unlike conventional approaches that minimize reconstruction loss independently at each layer, S2S trains sparse features at each intermediate layer to predict the sparse features of the immediately following layer, with the final layer anchored by an output-distribution loss and optimization proceeding sequentially from the output layer to the input layer.

\subsection{Problem Setting}

Consider a language model with $L$ layers. Let $\mathbf{h}^{(t)}\in\mathbb{R}^d$ denote the residual-stream state entering layer $t$ for $t\in\{0,\ldots,L-1\}$, and let $\mathbf{h}^{(L)}$ denote the final state after layer $L-1$. The model produces logits $\mathbf{y}=\mathbf{W}_u\mathbf{h}^{(L)}$ using $\mathbf{W}_u\in\mathbb{R}^{V\times d}$.
Our goal is to learn sparse feature representations $\mathbf{z}^{(t)} \in \mathbb{R}^k$ (where $k > d$) at each layer such that the features capture information causally relevant to the model's outputs.
The overcomplete dictionary ($k > d$) allows disentangling superimposed features in the dense hidden representations, while sparsity ensures that only a small subset of features are active for any given input.

\subsection{Sparse Feature Transcoder}

We define a sparse feature transcoder with layer-specific encoder and decoder weights.
The encoder maps the hidden representation at layer $t$ to a sparse feature vector using top-$K$ selection:
\begin{equation}
    \mathbf{z}^{(t)} = \text{TopK}\left(\mathbf{W}_{\text{enc}}^{(t)} \mathbf{h}^{(t)} + \mathbf{b}_{\text{enc}}^{(t)}, K\right),
\end{equation}
where $\mathbf{W}_{\text{enc}}^{(t)} \in \mathbb{R}^{k \times d}$ is the encoder weight matrix, $\mathbf{b}_{\text{enc}}^{(t)} \in \mathbb{R}^k$ is the encoder bias for layer $t$, and $\text{TopK}(\cdot, K)$ retains the $K$ largest signed pre-activation values (not the largest absolute values) while setting all others to zero. We select by signed value so that active features write along their own decoder directions rather than their negations, which keeps the meaning of each feature tied to one direction.
This provides explicit control over sparsity without requiring auxiliary loss terms.

The decoder reconstructs a hidden representation from the sparse features:
\begin{equation}
    \hat{\mathbf{h}}^{(t+1)} = \mathbf{W}_{\text{dec}}^{(t)} \mathbf{z}^{(t)} + \mathbf{b}_{\text{dec}}^{(t)},
\end{equation}
where $\mathbf{W}_{\text{dec}}^{(t)} \in \mathbb{R}^{d \times k}$ and $\mathbf{b}_{\text{dec}}^{(t)} \in \mathbb{R}^d$ are the decoder parameters for layer $t$.
Let $F_{t+1:L-1}$ denote the frozen suffix of language-model layers after the predicted state, with $F_{L:L-1}$ equal to the identity. We define $p_{\text{orig}}=\operatorname{softmax}(\mathbf{y})$ and
\begin{equation}
    p_{\text{sparse}}^{(t)}=\operatorname{softmax}\!\left(\mathbf{W}_u F_{t+1:L-1}(\hat{\mathbf{h}}^{(t+1)})\right),
\end{equation}
so an intermediate prediction is propagated through the remaining frozen model before unembedding. At the final layer this reduces to direct unembedding.

\begin{figure}[t!]
\centering
\includegraphics[width=\columnwidth]{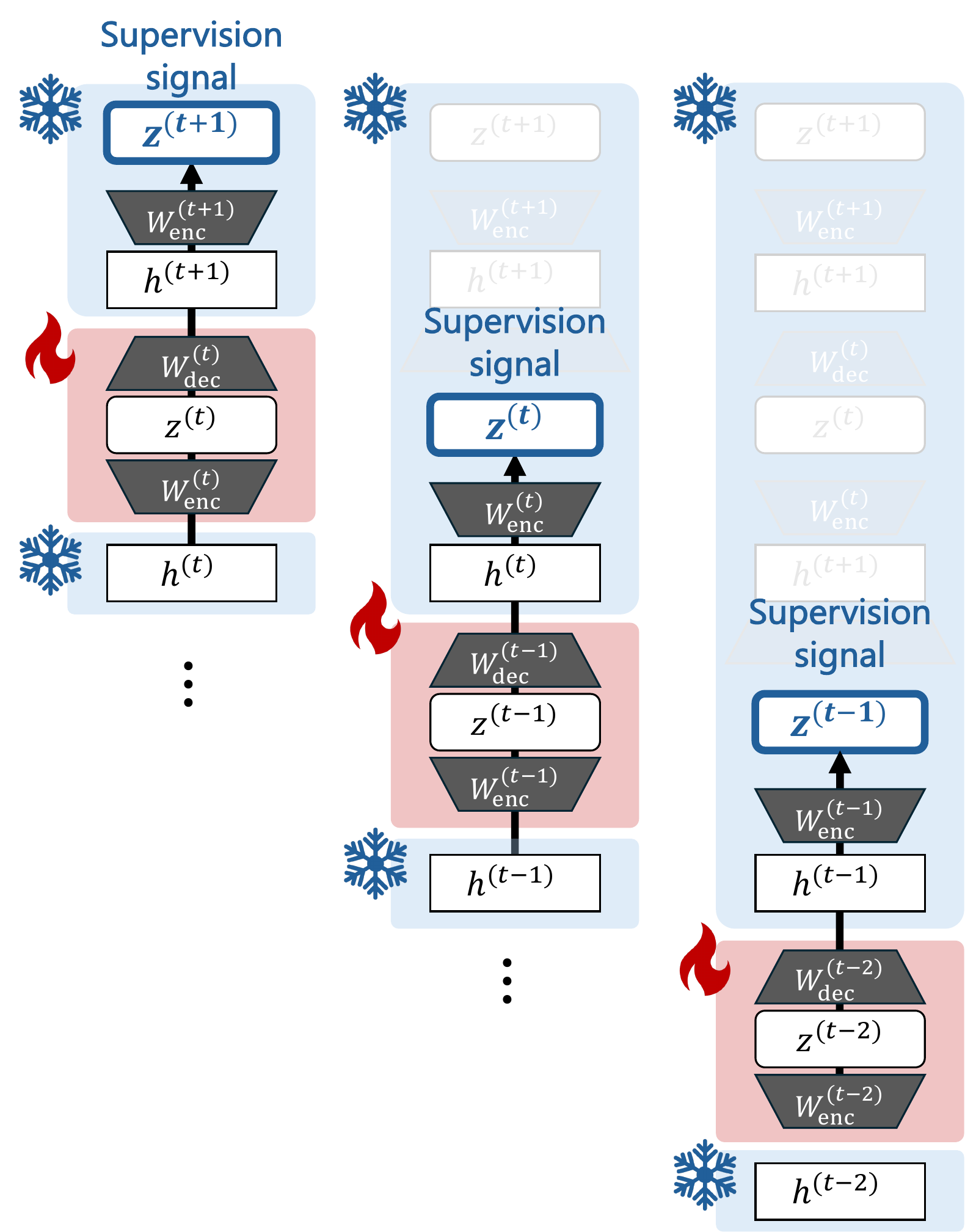}
\caption{Overview of S2S's backward chaining fine-tuning (Phase 2). Training proceeds sequentially from output to input layers. At each step, the transcoder at layer $t$ (shown in red with flame icons) is trained to predict the sparse features $\mathbf{z}^{(t+1)}$ of the next layer, while previously trained layers (shown in blue with snowflake icons) remain frozen.}
\label{fig:method}
\end{figure}

\subsection{Two-Phase Training}

S2S employs a two-phase training procedure: pretraining followed by backward chaining fine-tuning.
The pretraining phase stabilizes learning by first teaching the transcoder to predict next-layer hidden representations.
However, if we used the same sparsity level in both phases, the transcoder could simply reproduce the pretrained $\hat{\mathbf{h}}^{(t+1)}$ without further refinement.
Moreover, our motivation for using $\mathbf{z}^{(t+1)}$ as the prediction target stems from the goal of learning explicit correspondences between sparse features across layers, rather than mapping to dense representations that may contain redundant or entangled information.
By applying looser sparsity ($K_{\text{pre}}$) when predicting $\mathbf{h}^{(t+1)}$ and stricter sparsity ($K_{\text{fine}} < K_{\text{pre}}$) when predicting $\mathbf{z}^{(t+1)}$, we progressively refine inter-layer feature correspondences.

\paragraph{Phase 1: Pretraining.}
We first pretrain the layer-specific transcoders to predict target representations across all layers using a larger number of active features $K_{\text{pre}}$.
For intermediate layers, the transcoder learns to predict the next-layer hidden representations:
\begin{equation}
    \mathcal{L}_{\text{inter}}^{(t)} = \left\| \hat{\mathbf{h}}^{(t+1)} - \mathbf{h}^{(t+1)} \right\|_2^2.
\end{equation}
For the final layer, the transcoder learns to predict output logits through the frozen unembedding matrix:
\begin{equation}
    \mathcal{L}_{\text{final}}^{(L-1)} = D_{\text{KL}}\left( p_{\text{orig}} \| p_{\text{sparse}}^{(L-1)} \right).
    \label{eq:final_kl}
\end{equation}
This phase initializes the feature directions to capture meaningful transformations, while the looser sparsity constraint ($K_{\text{pre}}$) allows broad information retention.

\paragraph{Phase 2: Backward Chaining Fine-tuning.}
After pretraining, we fine-tune the transcoders using backward chaining with stricter sparsity constraints ($K_{\text{fine}} < K_{\text{pre}}$), as illustrated in \autoref{fig:method}.
The key insight is that sparse features should be learned in a backward manner, starting from the output layer and proceeding toward the input layer.

For the final layer $L-1$, we minimize the same output-distribution loss as in Phase~1 (\autoref{eq:final_kl}), now computed under the stricter sparsity budget $K_{\text{fine}}$.

For intermediate layers, we minimize the prediction error on the sparse features of the next layer:
\begin{align}
\hat{\mathbf{a}}^{(t+1)} &=
\mathbf{W}_{\mathrm{enc}}^{(t+1)}\hat{\mathbf{h}}^{(t+1)}
+ \mathbf{b}_{\mathrm{enc}}^{(t+1)}, \\
\hat{\mathbf{z}}^{(t+1)} &=
\operatorname{TopK}\!\left(\hat{\mathbf{a}}^{(t+1)},K_{\mathrm{fine}}\right), \\
\mathcal{L}^{(t)}_{\text{inter}} &=
\left\|\hat{\mathbf{z}}^{(t+1)}-\mathbf{z}^{(t+1)}\right\|_2^2,
\end{align}
where the target is computed from the actual hidden representation with the same sparsity budget, $\mathbf{z}^{(t+1)} = \operatorname{TopK}\!\left(\mathbf{W}_{\mathrm{enc}}^{(t+1)} \mathbf{h}^{(t+1)} + \mathbf{b}_{\mathrm{enc}}^{(t+1)},\, K_{\mathrm{fine}}\right)$, and treated as fixed (no gradients flow into the layer-$(t{+}1)$ transcoder).

The name \emph{sparse-to-sparse} refers to this Phase~2 objective, not to the type of the original language-model state. The layer-$t$ encoder receives dense $\mathbf{h}^{(t)}$ and produces sparse $\mathbf{z}^{(t)}$; after decoding through the model-aligned dense interface, its prediction is re-encoded and TopK-selected before comparison with the sparse target $\mathbf{z}^{(t+1)}$. Thus both the learned source representation and the supervised target are sparse, even though the frozen model's computational interface remains dense.
Appendix~\ref{sec:direct_map_appendix} compares this design with identity and direct sparse-to-sparse mapping controls.

The stricter sparsity constraint in Phase 2 ($K_{\text{fine}}$) encourages the model to concentrate adjacent-code prediction in a smaller active set. This design may suppress features that contribute less to the Phase~2 objective, but the sparsity schedule alone does not establish that every discarded feature is output-irrelevant.
This creates a chain of progressively constrained sparse representations from the final layer back to the input layer.

\section{Experimental Setup}
\label{sec:experimental_setup}

\subsection{Models and Training}

\paragraph{Models.}
We evaluate on eight language models spanning different scales and architectures: GPT-2 Small (124M), GPT-2 Medium (355M), GPT-2 Large (774M)~\citep{radford2019language}, Pythia-160M, Pythia-410M, Pythia-1B~\citep{biderman2023pythia}, Gemma-2-2B~\citep{team2024gemma2}, and Llama-3.2-1B~\citep{grattafiori2024llama}, following the model selection in prior SAE research~\citep{gao2025scaling, lieberum2024gemma, he2024llama}.

\paragraph{Training Data.}
Each baseline makes one pass over a 1B-token OpenWebText corpus~\citep{Gokaslan2019OpenWeb}. S2S uses exactly the same 1B unique tokens, making one pass in Phase~1 and a second in Phase~2. Thus it differs only in optimization passes and objectives, not unique training data. For GPT-2 Large, each adjacent transition receives about 27.8M tokens and 6,781 optimizer updates in Phase~2.
We use a disjoint 10K-token OpenWebText model-selection set that is not included in training.
Hidden representations are extracted from each layer using a context length of 512 tokens.

Downstream feature identification, strength calibration, and testing use disjoint data. For SST-2, these are the official train/dev/test splits (67,349/872/1,821); for TruthfulQA, we use a fixed category-stratified 491/163/163 split of its 817 questions. Split indices are shared across methods. All tunable choices are made without access to test results: sparse-model training hyperparameters are selected on the OpenWebText model-selection set, while task-specific steering strengths, matched-KL budgets, and other downstream calibration choices are selected exclusively on the corresponding development split. Test splits are used only once the full procedure and all settings have been fixed, for reporting final results.

\paragraph{Hyperparameters.}
We set the dictionary size $k = 16d$ for all models, following prior work~\citep{gao2025scaling}.
For Phase 1 (pretraining), we use a learning rate of $10^{-3}$ and $K_{\text{pre}} = 64$ active features.
For Phase 2 (backward chaining fine-tuning), we use a learning rate of $10^{-4}$ and $K_{\text{fine}} = 32$ active features.
We use a batch size of 4096 tokens for all experiments.
Training hyperparameters were selected on the 10K-token OpenWebText model-selection set.
We explored learning rates in $\{10^{-4}, 5 \times 10^{-4}, 10^{-3}, 5 \times 10^{-3}\}$, $K_{\text{pre}} \in \{32, 64, 128\}$ and $K_{\text{fine}} \in \{16, 32, 64\}$, and dictionary sizes $k \in \{4d, 8d, 16d, 32d\}$.
The chosen configuration achieved the best output KL divergence on that set.
We used NVIDIA A100 80GB GPUs for our experiments. S2S takes about 1.5 times as long to train as an SAE and well under half as long as e2e-SAE, and Appendix~\ref{sec:compute_cost} gives the full costs.
All results are averaged over five random seeds shared across methods.

\paragraph{Optimization.}
We use Adam optimizer~\citep{kingma2014adam} with cosine annealing learning rate schedule.
Phase 1 uses a learning rate of $10^{-3}$ for rapid convergence, while Phase 2 uses $10^{-4}$ to preserve pretrained feature directions while adjusting for output relevance.
Gradient clipping with a maximum norm of 1.0 is applied to stabilize training.

\paragraph{Decoder Normalization.}
For the TopK-based methods, following prior work~\citep{bricken2023monosemanticity,gao2025scaling}, we normalize each decoder column to have unit norm after each optimization step:
\begin{equation}
    \mathbf{W}_{\text{dec}}^{(t)}[:, i] \leftarrow \frac{\mathbf{W}_{\text{dec}}^{(t)}[:, i]}{\|\mathbf{W}_{\text{dec}}^{(t)}[:, i]\|_2},
\end{equation}
where $i$ indexes each feature.
This normalization makes signed activation values comparable across decoder directions, making the top-$K$ selection more meaningful and preventing degenerate solutions where some decoder directions dominate. CLT instead retains its original decoder-norm-weighted sparsity penalty, described below.

\subsection{Baselines}

We compare S2S against seven representative methods: SAE~\citep{bricken2023monosemanticity}, Transcoder~\citep{dunefsky2024transcoders}, e2e-SAE~\citep{braun2024e2e}, CLT~\citep{ameisen2025circuit}, Crosscoders~\citep{lindsey2024crosscoders}, TopK-SAE~\citep{gao2025scaling}, and a Shared Encoder variant. We reimplemented all methods in one codebase. For the TopK-based baselines, we use dictionary size $k=16d$ per layer and match the active-feature budget, token data, batch size, and optimizer search space. Crosscoders retain their standard multi-layer read/write structure. Shared Encoder uses a common $\mathbf{W}_{\text{enc}}$ with layer-specific decoders and tests hard parameter sharing.

\paragraph{CLT configuration.}
We follow the standard jointly trained multi-output CLT architecture and objective~\citep{ameisen2025circuit}. A JumpReLU feature $\mathbf{a}^{(s)}$ reads the residual stream at layer $s$ and has a separate decoder for every MLP output at layer $\ell\geq s$, giving $\hat{\mathbf{y}}^{(\ell)}=\sum_{s\leq\ell}\mathbf{W}_{\mathrm{dec}}^{s\rightarrow\ell}\mathbf{a}^{(s)}$. We jointly minimize reconstruction error summed over all MLP layers plus the original decoder-norm-weighted sparsity penalty. CLT uses the same per-layer width ($16d$), training tokens, batch size, and learning-rate search as the other methods. Its threshold and sparsity coefficient are selected on the OpenWebText model-selection set to match S2S's mean active-feature count within 5\%, with reconstruction error breaking ties. Because its multi-output decoders add parameters, we do not claim parameter matching and report wall-clock cost as the practical efficiency comparison. All downstream metrics use the same held-out examples and evaluation code.

\subsection{Evaluation Metrics}
\label{sec:evaluation_metrics}

We organize our evaluation along the three complementary dimensions outlined in the introduction: (1) inter-layer feature correspondence, (2) intervention effectiveness, and (3) output preservation.

\subsubsection{Inter-layer Feature Correspondence}
\label{sec:metrics_correspondence}

To directly assess whether S2S aligns sparse features across layers, we analyze the structure of the feature correspondence matrix and statistical relationships between pre-activations.

For any method with layer-specific encoders and decoders, we define the \textbf{feature correspondence matrix}:
\begin{equation}
    \mathbf{M}^{(t)} = \mathbf{W}_{\text{enc}}^{(t+1)} \mathbf{W}_{\text{dec}}^{(t)} \in \mathbb{R}^{k \times k},
\end{equation}
where $M^{(t)}_{ji}$ is a linearized decoder-to-encoder coupling from feature $i$ at layer $t$ to feature $j$ at layer $t+1$. Because this matrix omits biases and the input-dependent TopK support, it summarizes potential linear coupling rather than an exact causal contribution.

To measure coupling concentration independently of sign, we normalize the absolute entries within each source-feature column:
\begin{equation}
    p_{ji}^{(t)} = \frac{|M_{ji}^{(t)}|}{\sum_{j'=1}^{k}|M_{j'i}^{(t)}| + \epsilon}.
\end{equation}
We verified that no source-feature column had zero $\ell_1$ norm in any evaluated model or random seed. Thus, the $\epsilon$ term serves only as a numerical safeguard and does not determine the reported concentration values.
We define the \textbf{Top-5 Ratio} as the average probability mass assigned to the five largest entries in each column:
\begin{equation}
    \operatorname{Top5}^{(t)} = \frac{1}{k}\sum_{i=1}^{k}
    \sum_{j\in\operatorname{Top5}(p_{\cdot i}^{(t)})}p_{ji}^{(t)}.
\end{equation}
We define the \textbf{column-wise effective rank} as the average exponential entropy of the normalized columns:
\begin{align}
    H_i^{(t)} &= -\sum_{j=1}^{k}p_{ji}^{(t)}\log\bigl(p_{ji}^{(t)}+\epsilon\bigr), \\
    R_{\mathrm{eff}}^{(t)} &= \frac{1}{k}\sum_{i=1}^{k}\exp\bigl(H_i^{(t)}\bigr).
\end{align}
Top-5 Ratio summarizes the fixed-size head, whereas column-wise effective rank summarizes the full distribution. Higher Top-5 and lower effective rank indicate more concentrated coupling. This quantity is an entropy-based effective support size, not the spectral effective rank of the full matrix.

Top-5 Ratio and column-wise effective rank are computed directly from the correspondence weights and averaged over source features and adjacent-layer pairs.
We also report pre-activation-based metrics in \autoref{sec:sparse_analysis_appendix}: linear predictability ($R^2$) measuring how well layer $t$'s pre-activations predict layer $t+1$'s, and Centered Kernel Alignment (CKA)~\citep{kornblith2019similarity} measuring similarity of pre-activation geometry up to orthogonal transformations and isotropic scaling.
Because S2S directly optimizes an adjacent-code prediction loss, all correspondence metrics are partly coupled to the objective. We consequently use them to verify that the optimization has its intended structural effect, not as stand-alone evidence of interpretability. Held-out interventions, output ablation, and non-SAE controls provide evaluations not directly optimized by the correspondence loss.

Random training seed is the unit of replication. Metrics are first aggregated within seed, and S2S is then compared with each baseline using a two-sided paired $t$-test over the five matched seeds. Within each metric, we apply Holm correction~\citep{holm1979simple} over the 56 comparisons (8 models $\times$ 7 baselines) and mark a result as significant only when S2S is better than every baseline at adjusted $p<0.01$.

\subsubsection{Intervention Effectiveness}
\label{sec:metrics_intervention}

To demonstrate the practical utility of learned representations, we evaluate their effectiveness in controlling model outputs on two attributes: sentiment and truthfulness.

\paragraph{Feature Identification.}
For each task, we identify task-relevant features using a contrastive approach following \citet{turner2023steering}.
The procedure requires positive and negative output sets, but does not require both labels for the same prompt. We use unpaired class-conditional sentence means for SST-2~\citep{socher-etal-2013-recursive} and question-matched correct/incorrect contrasts for TruthfulQA~\citep{lin-etal-2022-truthfulqa}. Appendix~\ref{sec:data_protocol} gives the exact construction, templates, and token aggregation.

We compute a sparse steering vector $\mathbf{v} \in \mathbb{R}^k$ where each component represents the relevance of feature $i$ to the target attribute:
\begin{equation}
    v_i = \frac{1}{|\mathcal{S}^+|} \sum_{(x, y) \in \mathcal{S}^+} \bar{z}_i(y) - \frac{1}{|\mathcal{S}^-|} \sum_{(x, y) \in \mathcal{S}^-} \bar{z}_i(y),
\end{equation}
where $(x, y)$ denotes a prompt-output pair, and $\bar{z}_i(y)$ is the mean activation of feature $i$ over the output tokens $y$.

\paragraph{Intervention Protocol.}
We adapt the activation addition method from \citet{turner2023steering} to sparse feature spaces.
During generation, we add the scaled steering vector to the sparse feature activations at each layer:
\begin{equation}
    \mathbf{z}' = \mathbf{z} + \alpha \cdot \mathbf{v},
\end{equation}
where $\alpha > 0$ steers toward the target attribute (positive sentiment or truthfulness) and $\alpha < 0$ steers away from it.
The intervened hidden state is then computed via the decoder:
\begin{equation}
    \hat{\mathbf{h}}' = \mathbf{W}_{\text{dec}} \mathbf{z}' + \mathbf{b}_{\text{dec}},
\end{equation}
and passed to subsequent layers.
We apply this intervention at all layers during autoregressive generation. Unlike the single-transition replacement in \autoref{tab:general_metrics}, this pathway introduces approximation error at every layer, and that error propagates to later layers.
Because nominal $\alpha$ can induce different output changes across methods, we calibrate it separately for each method, task, seed, and direction and compare methods at common pre-specified intervention-induced output-KL budgets. Task changes are measured from the method-matched $\alpha=0$ sparse pathway, while reconstruction shift from the original model to $\alpha=0$ is reported separately. Appendix~\ref{sec:data_protocol} defines the KL, calibration grid, interpolation, and controls.

\paragraph{Sentiment Shift (SS).}
We generate continuations of 50 tokens using nucleus sampling ($p = 0.9$) and measure the sentiment of the generated continuation (excluding the prompt) using DistilBERT fine-tuned on SST-2~\citep{sanh2019distilbert}.
We evaluate steering in both directions and report the average absolute shift in positive class probability:
\begin{equation}
    \text{SS} = \frac{1}{2} \left( |\Delta p^{+}| + |\Delta p^{-}| \right),
\end{equation}
where $\Delta p^{+}$ and $\Delta p^{-}$ denote the change in $p_{\text{pos}}$ for positive and negative steering, respectively.
Each change is relative to the same method's $\alpha=0$ sparse-pathway output, using matched prompts and common random numbers.
Higher SS indicates more effective sentiment steering.

\paragraph{Truthfulness Rate (TF).}
Following \citet{li2023inference}, we evaluate on the multiple-choice task of TruthfulQA~\citep{lin-etal-2022-truthfulqa}.
We report the improvement in MC1 accuracy relative to the same method's $\alpha=0$ sparse-pathway control. Candidate answers are scored independently by their conditional completion log-probability following the question, matching the official MC1 protocol.
A greater improvement indicates more effective truthfulness enhancement.

\begin{table*}[t!]
\centering
\footnotesize
\resizebox{0.86\textwidth}{!}{%
\begin{tabular}{l|cc|cc|cc|cc}
\toprule
& \multicolumn{2}{c|}{GPT-2 Small} & \multicolumn{2}{c|}{GPT-2 Medium} & \multicolumn{2}{c|}{GPT-2 Large} & \multicolumn{2}{c}{Pythia-160M} \\
Method & Top-5 & Eff.\ Rank & Top-5 & Eff.\ Rank & Top-5 & Eff.\ Rank & Top-5 & Eff.\ Rank \\
\midrule
SAE & 0.206 & 51.19 & 0.212 & 48.41 & 0.231 & 45.09 & 0.195 & 52.81 \\
Transcoder & 0.268 & 35.41 & 0.287 & 32.79 & 0.295 & 30.21 & 0.268 & 36.69 \\
e2e-SAE & 0.298 & 31.19 & 0.305 & 29.11 & 0.327 & 26.79 & 0.286 & 32.41 \\
CLT & 0.315 & 28.41 & 0.334 & 26.19 & 0.344 & 24.11 & 0.315 & 29.49 \\
Crosscoders & \underline{0.345} & \underline{25.59} & \underline{0.352} & \underline{23.81} & \underline{0.374} & \underline{21.89} & \underline{0.333} & \underline{26.41} \\
TopK-SAE & 0.245 & 38.91 & 0.264 & 36.39 & 0.272 & 33.81 & 0.245 & 40.09 \\
Shared Enc. & 0.185 & 56.29 & 0.188 & 54.11 & 0.207 & 51.19 & 0.172 & 58.21 \\
\midrule
S2S & $\mathbf{0.495}^\dagger$ & $\mathbf{16.21}^\dagger$ & $\mathbf{0.494}^\dagger$ & $\mathbf{16.49}^\dagger$ & $\mathbf{0.545}^\dagger$ & $\mathbf{13.21}^\dagger$ & $\mathbf{0.488}^\dagger$ & $\mathbf{17.09}^\dagger$ \\
\bottomrule
\end{tabular}%
}
\vspace{0.3cm}
\resizebox{0.86\textwidth}{!}{%
\begin{tabular}{l|cc|cc|cc|cc}
\toprule
& \multicolumn{2}{c|}{Pythia-410M} & \multicolumn{2}{c|}{Pythia-1B} & \multicolumn{2}{c|}{Gemma-2-2B} & \multicolumn{2}{c}{Llama-3.2-1B} \\
Method & Top-5 & Eff.\ Rank & Top-5 & Eff.\ Rank & Top-5 & Eff.\ Rank & Top-5 & Eff.\ Rank \\
\midrule
SAE & 0.218 & 47.80 & 0.231 & 44.60 & 0.238 & 43.20 & 0.235 & 44.10 \\
Transcoder & 0.289 & 32.10 & 0.304 & 29.40 & 0.312 & 28.10 & 0.308 & 28.80 \\
e2e-SAE & 0.314 & 28.50 & 0.332 & 25.80 & 0.341 & 24.60 & 0.337 & 25.20 \\
CLT & 0.338 & 25.40 & 0.356 & 22.80 & 0.364 & 21.70 & 0.361 & 22.30 \\
Crosscoders & \underline{0.362} & \underline{23.10} & \underline{0.382} & \underline{20.40} & \underline{0.391} & \underline{19.50} & \underline{0.387} & \underline{19.90} \\
TopK-SAE & 0.267 & 35.20 & 0.283 & 32.10 & 0.291 & 30.80 & 0.287 & 31.40 \\
Shared Enc. & 0.198 & 53.40 & 0.212 & 50.10 & 0.219 & 48.50 & 0.215 & 49.20 \\
\midrule
S2S & $\mathbf{0.534}^\dagger$ & $\mathbf{14.40}^\dagger$ & $\mathbf{0.568}^\dagger$ & $\mathbf{12.40}^\dagger$ & $\mathbf{0.592}^\dagger$ & $\mathbf{11.20}^\dagger$ & $\mathbf{0.578}^\dagger$ & $\mathbf{11.80}^\dagger$ \\
\bottomrule
\end{tabular}%
}
\caption{Task-independent feature correspondence analysis across all models. Top-5: absolute-weight column concentration ratio ($\uparrow$). Eff.\ Rank: entropy-based column-wise effective rank ($\downarrow$). Best results are in \textbf{bold}, second best are \underline{underlined}. Random seed is the unit of replication. $\dagger$ denotes $p<0.01$ against every baseline (two-sided paired $t$-test, Holm-adjusted over 56 comparisons).}
\label{tab:matrix_analysis}
\end{table*}

\subsubsection{Output Preservation}
\label{sec:metrics_preservation}

To ensure that sparse features capture information retained by the model, we evaluate whether output distributions are preserved when decoding sparse representations back to dense vectors.
We compute these metrics on 10K tokens from OpenWebText, excluding data used during training.

\paragraph{Output KL Divergence ($D_{\text{KL}}^{\text{out}}$).}
Extending the approach of \citet{braun2024e2e}, we measure how well the learned sparse features preserve the model's output distribution.
For S2S and transcoder-style methods, at transition $t$ we encode $\mathbf{h}^{(t)}$ and use the decoded prediction $\hat{\mathbf{h}}^{(t+1)}$ in place of the original post-layer state $\mathbf{h}^{(t+1)}$. For same-layer autoencoder baselines such as SAE and TopK-SAE, we instead encode the actual post-layer state $\mathbf{h}^{(t+1)}$ with the autoencoder assigned to that state and replace $\mathbf{h}^{(t+1)}$ with its reconstruction. Thus every method is evaluated by replacing the same post-layer state before propagation through the remaining frozen model layers.
We compute:
\begin{align}
    D_{\text{KL}}^{\text{out}}(t) &= D_{\text{KL}}\left( p_{\text{orig}}(y) \| p_{\text{sparse}}^{(t)}(y) \right), \\
    \bar{D}_{\text{KL}}^{\text{out}} &= \frac{1}{L} \sum_{t=0}^{L-1} D_{\text{KL}}^{\text{out}}(t),
\end{align}
where $p_{\text{orig}}(y)$ is the original model's output distribution and $p_{\text{sparse}}^{(t)}(y)$ is the output distribution when the hidden state at layer $t$ passes through the sparse bottleneck.
Lower $\bar{D}_{\text{KL}}^{\text{out}}$ indicates that the sparse features retain output-relevant information.

\paragraph{Causal Importance (CI).}
To verify that high-activation features are causally important for outputs, we measure how much the output distribution changes when the $m$ active features with the largest signed coefficients are ablated (zeroed); selection is by value, not absolute value.
Specifically, we compute the increase in KL divergence between the original output distribution and the sparse pathway output after this ablation:
\begin{align}
    \text{CI}_m = \frac{1}{L} \sum_{t=0}^{L-1} \Bigl[{}
    &D_{\text{KL}}\left(p_{\text{orig}} \| \tilde{p}^{(t)}\right) \nonumber \\
    &{}-D_{\text{KL}}\left(p_{\text{orig}} \| p_{\text{sparse}}^{(t)}\right)\Bigr],
\end{align}
where $p_{\text{sparse}}^{(t)}$ is the output distribution when layer $t$'s hidden state passes through the sparse pathway, and $\tilde{p}^{(t)}$ is the output distribution after ablating the $m$ highest-activation features at layer $t$.
We report CI averaged over $m \in \{1, 2, 4, 8\}$.
Higher CI indicates that high-activation features capture more causally important information, as their removal causes larger distributional shifts in the model's output.

\section{Results}
\label{sec:results}

\begin{table*}[t!]
\centering
\footnotesize
\resizebox{0.86\textwidth}{!}{
\begin{tabular}{l|cc|cc|cc|cc}
\toprule
& \multicolumn{2}{c|}{GPT-2 Small} & \multicolumn{2}{c|}{GPT-2 Medium} & \multicolumn{2}{c|}{GPT-2 Large} & \multicolumn{2}{c}{Pythia-160M} \\
Method & SS & TF & SS & TF & SS & TF & SS & TF \\
\midrule
SAE & 0.124 & 0.203 & 0.139 & 0.274 & 0.146 & 0.293 & 0.115 & 0.186 \\
Transcoder & 0.131 & 0.234 & 0.148 & 0.302 & 0.155 & 0.332 & 0.122 & 0.214 \\
e2e-SAE & \underline{0.155} & \underline{0.312} & \underline{0.171} & \underline{0.403} & 0.174 & \underline{0.432} & \textbf{0.158} & \underline{0.283} \\
CLT & 0.146 & 0.284 & 0.160 & 0.362 & 0.170 & 0.391 & 0.137 & 0.253 \\
Crosscoders & 0.149 & 0.293 & 0.167 & 0.382 & \underline{0.178} & 0.412 & 0.140 & 0.264 \\
TopK-SAE & 0.138 & 0.253 & 0.156 & 0.333 & 0.162 & 0.352 & 0.129 & 0.233 \\
Shared Enc. & 0.118 & 0.192 & 0.127 & 0.248 & 0.133 & 0.267 & 0.108 & 0.172 \\
\midrule
S2S & $\mathbf{0.167}$ & $\mathbf{0.347}^\dagger$ & $\mathbf{0.186}$ & $\mathbf{0.458}^\dagger$ & $\mathbf{0.197}^\dagger$ & $\mathbf{0.498}^\dagger$ & \underline{0.152} & $\mathbf{0.308}$ \\
\bottomrule
\end{tabular}
}
\vspace{0.3cm}
\resizebox{0.86\textwidth}{!}{
\begin{tabular}{l|cc|cc|cc|cc}
\toprule
& \multicolumn{2}{c|}{Pythia-410M} & \multicolumn{2}{c|}{Pythia-1B} & \multicolumn{2}{c|}{Gemma-2-2B} & \multicolumn{2}{c}{Llama-3.2-1B} \\
Method & SS & TF & SS & TF & SS & TF & SS & TF \\
\midrule
SAE & 0.137 & 0.241 & 0.146 & 0.304 & 0.161 & 0.331 & 0.151 & 0.314 \\
Transcoder & 0.142 & 0.285 & 0.160 & 0.340 & 0.167 & 0.374 & 0.163 & 0.354 \\
e2e-SAE & \underline{0.168} & \underline{0.370} & \underline{0.178} & \underline{0.444} & \underline{0.193} & \underline{0.471} & \underline{0.191} & 0.453 \\
CLT & 0.156 & 0.335 & 0.174 & 0.400 & 0.181 & 0.434 & 0.178 & 0.420 \\
Crosscoders & 0.162 & 0.350 & 0.174 & 0.424 & 0.185 & 0.455 & 0.182 & \underline{0.464} \\
TopK-SAE & 0.149 & 0.305 & 0.166 & 0.360 & 0.177 & 0.400 & 0.170 & 0.384 \\
Shared Enc. & 0.123 & 0.216 & 0.129 & 0.265 & 0.140 & 0.300 & 0.136 & 0.284 \\
\midrule
S2S & $\mathbf{0.184}^\dagger$ & $\mathbf{0.416}^\dagger$ & $\mathbf{0.202}^\dagger$ & $\mathbf{0.506}^\dagger$ & $\mathbf{0.213}^\dagger$ & $\mathbf{0.550}^\dagger$ & $\mathbf{0.206}^\dagger$ & $\mathbf{0.521}^\dagger$ \\

\bottomrule
\end{tabular}
}
\caption{Intervention results at the primary matched output-KL budget, averaged over five seeds. SS: Sentiment Shift ($\uparrow$); TF: Truthfulness Rate improvement ($\uparrow$). Changes are relative to the method-matched $\alpha=0$ sparse pathway. Best results are in \textbf{bold}, second best are \underline{underlined}. $\dagger$ denotes $p<0.01$ against every baseline (two-sided paired $t$-test, Holm-adjusted over 56 comparisons).}
\label{tab:main_results}
\end{table*}

\subsection{Inter-layer Feature Correspondence}
\label{sec:results_correspondence}

\autoref{tab:matrix_analysis} shows that S2S has the highest Top-5 concentration and lowest column-wise effective rank on every model, and it is significantly better than every baseline throughout. Each source feature therefore maps to a focused target set rather than distributing its mass broadly. The pre-activation analyses in Appendix~\ref{sec:sparse_analysis_appendix} agree: S2S substantially improves both linear predictability and CKA between adjacent layers.

Shared Encoder is an informative counterexample: hard index sharing does not improve task-independent correspondence or output preservation. This shows that imposing shared feature indices is not sufficient to recover meaningful cross-layer relationships.

The correspondence gain comes from the sparse target rather than from the extra optimization pass. A control that starts from the same Phase~1 checkpoint and follows the same Phase~2 schedule, but keeps the dense reconstruction objective, stays far below S2S in correspondence, and the hidden-state reconstruction cost of S2S relative to this control is small. Appendix~\ref{sec:additional_analyses} reports this control together with an additional training pass for e2e-SAE and a single-seed S2S run on Mistral-7B~\citep{jiang2023mistral}, where Phase~2 again sharpens correspondence.

Chaining adjacent correspondences lets us follow individual features across several layers of GPT-2 Large. A feature that responds to positive evaluations such as ``excellent performance'' keeps a concentrated successor from layer 14 to layer 17, while other chains show a speech feature splitting into content and attribution, attribution features merging, and a local-syntax feature gradually losing its successor. Over 20 chains, S2S keeps a more concentrated successor than e2e-SAE at every hop. In a preliminary check where one LLM describes features and a second, method-blinded LLM judges whether a source feature and its strongest target express the same concept, S2S pairs are judged consistent more often than e2e-SAE pairs.

\subsection{Intervention Task Performance}
\label{sec:results_intervention}

\autoref{tab:main_results} reports performance at the primary matched output-KL budget.
S2S achieves the largest gains in 15 of 16 model-task settings, and it is significantly better than every baseline in 12 of them.
The exception is SS on Pythia-160M, where e2e-SAE is slightly stronger, and S2S's TF advantage on the same model does not reach significance. Pythia-160M is also the model where e2e-SAE has higher CI, which suggests that the smallest model may leave too little capacity for adjacent-code prediction, although we have not tested this explanation.
Because all methods are calibrated to the same intervention-induced output KL and gains are measured from each method's own $\alpha=0$ pathway, these differences reflect steering efficiency at a fixed output-distribution shift rather than differences in intervention magnitude or sparse-pathway reconstruction quality. The ranking is unchanged at the 0.01 and 0.10 budgets. Since SS scores free 50-token continuations, it measures steering of open-ended generation rather than of a classification decision.

\subsection{Output Preservation}
\label{sec:results_preservation}

As shown in \autoref{tab:general_metrics}, across the eight models, S2S averages 0.048 output KL and 0.216 CI, compared with 0.052 and 0.185 for e2e-SAE, indicating stronger distribution preservation and causal relevance overall. There are two model-metric exceptions: e2e-SAE has lower KL on Pythia-410M (0.053 vs.\ 0.056), and higher CI on Pythia-160M (0.179 vs.\ 0.168). Reconstruction-only baselines otherwise have higher KL and lower CI, while Shared Encoder is worst on both measures, showing that shared-index alignment does not ensure output relevance.

Two further checks support these results. Ablating the top-active features changes the output more than ablating features matched in activation magnitude, and this margin is larger for S2S than for e2e-SAE. When all transcoders are inserted at once without steering, which is the reference pathway for \autoref{tab:main_results}, S2S also degrades perplexity and LAMBADA~\citep{paperno-etal-2016-lambada} accuracy less than Transcoder and the compute-matched dense control. Appendix~\ref{sec:additional_analyses} gives both analyses.

Because CI measures the distributional effect of feature ablation without task classifiers, it provides a task-independent check against evaluator-specific gains. It remains an aggregate diagnostic, not a complete circuit-level account.

\section{Related Work}

SAEs decompose dense representations into sparse features, but reconstruction-trained features need not be causally useful for steering~\citep{bricken2023monosemanticity,cunningham2023sparse,templeton2024scaling}. Feature-selection methods target more output-relevant or less disruptive interventions~\citep{arad2025saes,chalnev2024sae}. Recent benchmarking work also highlights the importance of evaluating steering methods using behavioral metrics and carefully matched intervention settings~\citep{wu2025axbench}. We therefore evaluate S2S at matched output-distribution shift and distinguish steering effectiveness from sparse-pathway reconstruction quality.

Transcoders, CLTs, Crosscoders, and e2e-SAEs extend sparse feature learning beyond independent reconstruction~\citep{dunefsky2024transcoders,ameisen2025circuit,lindsey2024crosscoders,braun2024e2e}. Behavior-specific analyses also trace how information is integrated across layers, for example how temporal anchors enter factual recall~\citep{niwa2027tracing}. S2S does not claim the general idea of cross-layer analysis. Its contribution is to predict the next layer's sparse code explicitly and propagate that supervision through a two-stage backward schedule.

\section{Conclusion}

We introduced S2S, which trains each layer's sparse code to predict the next using a two-stage backward schedule. Across eight models, S2S improves adjacent-layer structure, output preservation, and causal ablation relative to matched sparse-feature baselines.
At matched output-KL budgets, S2S leads in 15 of 16 model-task settings, supporting its use for cross-layer analysis and for steering sentiment and truthfulness.

\section*{Limitations}

Our structural correspondence metrics partially overlap with the Phase~2 objective and therefore cannot independently establish interpretability. Causal ablation reduces this circularity but remains an aggregate diagnostic. We do not yet provide blinded human studies, expert judgments of feature labels, or complete causal traces showing how one semantic feature transforms across every layer.

Matching intervention-induced output KL controls the magnitude of output-distribution change but does not guarantee equal semantic specificity or equal residual-stream geometry across methods. Our steering evaluation is limited to sentiment and truthfulness in English and does not establish robustness to adversarial prompts, long-form generation, or distribution shift.

Finally, TopK and a uniform active-feature budget stabilize backward chaining but may underallocate capacity at later layers. Appendix~\ref{sec:design_appendix} studies alternative sparsity levels, adaptive budgets, a joint loss, and BatchTopK/JumpReLU; adapting dynamic-sparsity objectives to S2S remains open.

\section*{Ethical Considerations}

S2S is a dual-use interpretability and steering method. Features that alter sentiment or improve truthfulness can support model analysis, but the same procedure could be used to identify and amplify harmful, biased, or deceptive behavior. Researchers deploying feature-steering systems should therefore apply appropriate access controls and monitor both intended and unintended behavioral changes.

\bibliography{references}

\clearpage
\appendix

\section{Dataset Construction and Matched-KL Protocol}
\label{sec:data_protocol}

Feature identification requires attribute-labeled outputs, not paired labels for a common prompt. For an unpaired dataset, the steering vector is the difference between class-conditional mean activations. When the dataset supplies matched alternatives, a within-input contrast reduces content confounding. Feature-identification, calibration, and test partitions are disjoint, and their example identifiers are shared across methods.

\paragraph{SST-2.}
We use the official train/dev/test splits (67,349/872/1,821). Feature identification uses complete labeled training sentences: $\mathcal{S}^{+}$ and $\mathcal{S}^{-}$ contain positive and negative sentences, respectively, and $\bar{\mathbf z}(y)$ averages over all non-special sentence tokens. We do not construct artificial counterfactual continuations. Teacher-forced development sentences are used exclusively to tune the steering strength and all matched-KL calibration choices. After these choices are fixed, each test sentence is supplied as a prompt, the model generates 50 tokens, and the sentiment classifier is applied only to the continuation. Positive and negative steering use the same test prompts and common random numbers. No test generations or scores are used for tuning or model selection.

\paragraph{TruthfulQA.}
We use a fixed category-stratified 491/163/163 split. Each candidate is formatted as \texttt{Question: \{question\}\textbackslash nAnswer: \{answer\}}. For question $q$, we compute
\begin{equation}
 \mathbf v_q=\bar{\mathbf z}(a_q^+)-
 \frac{1}{|A_q^-|}\sum_{a\in A_q^-}\bar{\mathbf z}(a),
\end{equation}
where $a_q^+$ is the reference best answer, $A_q^-$ contains that question's incorrect answers, and each activation mean uses answer tokens only. The steering vector averages $\mathbf v_q$ over training questions, so every question has equal weight. Development questions and candidates are used exclusively to tune the steering strength and all matched-KL calibration choices. After these choices are fixed, original MC1 candidates in the test split are scored independently by conditional completion log-probability, and the highest-scoring candidate is selected. No test questions, candidates, or scores are used for tuning or model selection. The newer binary-choice version is an optional robustness check rather than being mixed with original MC1.

\paragraph{Matched-KL calibration.}
For method $m$, fixed teacher-forced context $c$, and strength $\alpha$, let $p_{m,0}(\cdot\mid c)$ be the next-token distribution through the same sparse pathway with $\alpha=0$ and $p_{m,\alpha}(\cdot\mid c)$ the distribution under steering. We define
\begin{equation}
 D_m(\alpha)=\mathbb{E}_{c\in\mathcal{D}_{\mathrm{cal}}}
 D_{\mathrm{KL}}\!\left(p_{m,0}(\cdot\mid c)\,\|\,p_{m,\alpha}(\cdot\mid c)\right).
\end{equation}
For each method, task, seed, and direction, we evaluate a logarithmic strength grid on fixed development contexts. The endpoints are expanded until they bracket every common target budget or reach a pre-specified numerical-stability limit. We interpolate linearly in $\log\alpha$ between neighboring strengths to reach target $\kappa$. The shared budget set $\mathcal{K}=\{0.01, 0.05, 0.10\}$ nats/token and its primary middle budget are fixed before examining test scores, using only the range jointly supported by all methods. The main results use the primary budget of $0.05$ nats/token. Sparse-pathway reconstruction KL is reported separately in \autoref{tab:general_metrics}.

\begin{table*}[t!]
\centering
\footnotesize
\begin{tabular}{l|cc|cc|cc|cc}
\toprule
& \multicolumn{2}{c|}{GPT-2 Small} & \multicolumn{2}{c|}{GPT-2 Medium} & \multicolumn{2}{c|}{GPT-2 Large} & \multicolumn{2}{c}{Pythia-160M} \\
Method & $D_{\text{KL}}$ & CI & $D_{\text{KL}}$ & CI & $D_{\text{KL}}$ & CI & $D_{\text{KL}}$ & CI \\
\midrule
SAE & 0.094 & 0.076 & 0.091 & 0.083 & 0.088 & 0.089 & 0.097 & 0.070 \\
Transcoder & 0.081 & 0.109 & 0.078 & 0.118 & 0.075 & 0.125 & 0.084 & 0.102 \\
e2e-SAE & \underline{0.058} & \underline{0.164} & \underline{0.054} & \underline{0.178} & \underline{0.049} & \underline{0.191} & \underline{0.059} & \textbf{0.179} \\
CLT & 0.073 & 0.139 & 0.070 & 0.150 & 0.067 & 0.161 & 0.076 & 0.129 \\
Crosscoders & 0.066 & 0.152 & 0.065 & 0.162 & 0.060 & 0.175 & 0.069 & 0.142 \\
TopK-SAE & 0.087 & 0.095 & 0.084 & 0.105 & 0.081 & 0.112 & 0.090 & 0.089 \\
Shared Enc. & 0.102 & 0.068 & 0.098 & 0.074 & 0.095 & 0.079 & 0.105 & 0.062 \\
\midrule
S2S & $\mathbf{0.056}$ & $\mathbf{0.192}^\dagger$ & $\mathbf{0.050}$ & $\mathbf{0.212}^\dagger$ & $\mathbf{0.046}$ & $\mathbf{0.229}^\dagger$ & $\mathbf{0.051}^\dagger$ & \underline{0.168} \\
\bottomrule
\end{tabular}

\vspace{0.3cm}

\begin{tabular}{l|cc|cc|cc|cc}
\toprule
& \multicolumn{2}{c|}{Pythia-410M} & \multicolumn{2}{c|}{Pythia-1B} & \multicolumn{2}{c|}{Gemma-2-2B} & \multicolumn{2}{c}{Llama-3.2-1B} \\
Method & $D_{\text{KL}}$ & CI & $D_{\text{KL}}$ & CI & $D_{\text{KL}}$ & CI & $D_{\text{KL}}$ & CI \\
\midrule
SAE & 0.090 & 0.080 & 0.086 & 0.093 & 0.083 & 0.100 & 0.083 & 0.096 \\
Transcoder & 0.077 & 0.115 & 0.073 & 0.132 & 0.070 & 0.142 & 0.074 & 0.135 \\
e2e-SAE & \textbf{0.053} & \underline{0.172} & \underline{0.048} & \underline{0.195} & \underline{0.046} & \underline{0.205} & \underline{0.049} & \underline{0.199} \\
CLT & 0.069 & 0.145 & 0.065 & 0.165 & 0.062 & 0.172 & 0.064 & 0.169 \\
Crosscoders & 0.062 & 0.159 & 0.058 & 0.179 & 0.057 & 0.186 & 0.059 & 0.182 \\
TopK-SAE & 0.083 & 0.102 & 0.079 & 0.115 & 0.076 & 0.122 & 0.080 & 0.119 \\
Shared Enc. & 0.099 & 0.071 & 0.094 & 0.082 & 0.091 & 0.088 & 0.095 & 0.085 \\
\midrule
S2S & \underline{0.056} & $\mathbf{0.202}^\dagger$ & $\mathbf{0.041}^\dagger$ & $\mathbf{0.235}^\dagger$ & $\mathbf{0.041}$ & $\mathbf{0.251}^\dagger$ & $\mathbf{0.041}^\dagger$ & $\mathbf{0.242}^\dagger$ \\
\bottomrule
\end{tabular}
\caption{Output preservation metrics across all models. $D_{\text{KL}}^{\text{out}}$: Output KL Divergence ($\downarrow$), CI: Causal Importance ($\uparrow$). Best results are in \textbf{bold}, second best are \underline{underlined}. $\dagger$ denotes $p<0.01$ against every baseline (two-sided paired $t$-test, Holm-adjusted over 56 comparisons).}
\label{tab:general_metrics}
\end{table*}

\section{Inter-layer Feature Relationship Analysis}
\label{sec:sparse_analysis_appendix}

We analyze the statistical relationships between pre-activations to evaluate whether S2S learns meaningful inter-layer feature correspondences.

A key challenge is that sparse vectors $\mathbf{z}^{(t)}$ (after top-$K$ selection) have most dimensions set to zero, making standard metrics unreliable.
We therefore evaluate on pre-activation values before top-$K$ selection:
\begin{equation}
    \mathbf{a}^{(t)} = \mathbf{W}_{\text{enc}}^{(t)} \mathbf{h}^{(t)} + \mathbf{b}_{\text{enc}}^{(t)},
\end{equation}
which are dense vectors representing feature importance scores.
These dense score vectors allow us to evaluate whether adjacent layers exhibit predictable pre-selection structure.

We measure \textbf{linear predictability ($R^2$)} by learning a ridge-regression mapping $\mathbf{W}: \mathbf{a}^{(t)} \rightarrow \mathbf{a}^{(t+1)}$. We minimize
\begin{equation}
    \frac{1}{N_{\mathrm{fit}}}
    \left\|\mathbf{A}^{(t)}\mathbf{W}-\mathbf{A}^{(t+1)}\right\|_F^2
    + \lambda\|\mathbf{W}\|_F^2.
\end{equation}
We fit each mapping on 100K OpenWebText tokens, select $\lambda \in \{10^{-6},10^{-5},\ldots,10^{2}\}$ on a disjoint 10K-token development set, and report $R^2$ on a separate 10K-token test set. All three splits exclude sparse-feature training and model-selection data, and token indices are shared across methods. Pre-activations are standardized using only the fit-set statistics, which are then applied unchanged to the development and test sets.

Because the full $k\times k$ coefficient matrix is large, we solve the ridge objective with a matrix-free block LSMR procedure. Target features are partitioned into blocks of 256 dimensions. Each block is optimized to a relative tolerance of $10^{-4}$ with at most 500 iterations, evaluated on the test set, and discarded before the next block is processed. We aggregate the residual and total sums of squares across all target dimensions before computing the reported $R^2$, rather than averaging block-level scores. This procedure avoids materializing the full coefficient matrix while preserving the stated full-rank ridge objective. Higher $R^2$ indicates that feature importance scores at layer $t$ can predict those at layer $t+1$ on held-out data.

We compute \textbf{CKA (Centered Kernel Alignment)}~\citep{kornblith2019similarity} to measure structural similarity that is invariant to orthogonal transformations and isotropic scaling, but not to arbitrary invertible linear transformations.
Given $N$ samples, let $\mathbf{A}^{(t)}, \mathbf{A}^{(t+1)} \in \mathbb{R}^{N \times k}$ denote the pre-activation matrices.
We compute linear kernel matrices and calculate:
\begin{equation}
    \text{CKA} = \frac{\text{HSIC}(\mathbf{K}, \mathbf{L})}{\sqrt{\text{HSIC}(\mathbf{K}, \mathbf{K}) \cdot \text{HSIC}(\mathbf{L}, \mathbf{L})}},
\end{equation}
where HSIC is the Hilbert-Schmidt Independence Criterion.

\autoref{tab:sparse_analysis} shows that S2S achieves substantially higher $R^2$ and CKA compared to all baselines.
The large gap in $R^2$ indicates that S2S learns feature representations where layer $t$'s importance scores are highly predictive of layer $t+1$'s scores.
Similarly, the higher CKA indicates that S2S captures more consistent pre-activation geometry across adjacent layers.

Notably, Shared Encoder achieves the lowest $R^2$ and CKA scores despite using identical encoder weights across layers.
This counterintuitive result reveals that parameter sharing alone does not guarantee meaningful inter-layer feature correspondence.
Because the shared encoder must compromise between the distinct computational requirements of different layers, it learns a suboptimal representation that poorly captures the actual relationships between layer-specific hidden states.

\begin{table*}[t!]
\centering
\footnotesize
\begin{tabular}{l|cc|cc|cc|cc}
\toprule
& \multicolumn{2}{c|}{GPT-2 Small} & \multicolumn{2}{c|}{GPT-2 Medium} & \multicolumn{2}{c|}{GPT-2 Large} & \multicolumn{2}{c}{Pythia-160M} \\
Method & $R^2$ & CKA & $R^2$ & CKA & $R^2$ & CKA & $R^2$ & CKA \\
\midrule
SAE & 0.294 & 0.406 & 0.304 & 0.414 & 0.313 & 0.430 & 0.287 & 0.399 \\
Transcoder & 0.333 & 0.446 & 0.349 & 0.452 & 0.361 & 0.462 & 0.327 & 0.440 \\
e2e-SAE & 0.327 & 0.429 & 0.333 & 0.446 & 0.345 & 0.456 & 0.319 & 0.422 \\
CLT & 0.370 & 0.480 & 0.384 & 0.486 & 0.397 & 0.496 & 0.360 & 0.470 \\
Crosscoders & \underline{0.394} & \underline{0.496} & \underline{0.405} & \underline{0.513} & \underline{0.424} & \underline{0.522} & \underline{0.383} & \underline{0.493} \\
TopK-SAE & 0.313 & 0.426 & 0.327 & 0.434 & 0.335 & 0.450 & 0.309 & 0.414 \\
Shared Enc. & 0.270 & 0.380 & 0.273 & 0.390 & 0.286 & 0.394 & 0.259 & 0.377 \\
\midrule
S2S & $\mathbf{0.543}^\dagger$ & $\mathbf{0.650}^\dagger$ & $\mathbf{0.574}^\dagger$ & $\mathbf{0.669}^\dagger$ & $\mathbf{0.600}^\dagger$ & $\mathbf{0.700}^\dagger$ & $\mathbf{0.534}^\dagger$ & $\mathbf{0.632}^\dagger$ \\
\bottomrule
\end{tabular}

\vspace{0.3cm}

\begin{tabular}{l|cc|cc|cc|cc}
\toprule
& \multicolumn{2}{c|}{Pythia-410M} & \multicolumn{2}{c|}{Pythia-1B} & \multicolumn{2}{c|}{Gemma-2-2B} & \multicolumn{2}{c}{Llama-3.2-1B} \\
Method & $R^2$ & CKA & $R^2$ & CKA & $R^2$ & CKA & $R^2$ & CKA \\
\midrule
SAE & 0.307 & 0.419 & 0.315 & 0.433 & 0.324 & 0.436 & 0.325 & 0.446 \\
Transcoder & 0.350 & 0.460 & 0.367 & 0.469 & 0.371 & 0.472 & 0.373 & 0.483 \\
e2e-SAE & 0.341 & 0.444 & 0.350 & 0.460 & 0.357 & 0.459 & 0.361 & 0.464 \\
CLT & 0.385 & 0.493 & 0.404 & 0.502 & 0.403 & 0.510 & 0.411 & 0.509 \\
Crosscoders & \underline{0.414} & \underline{0.512} & \underline{0.427} & \underline{0.536} & \underline{0.434} & \underline{0.536} & \underline{0.437} & \underline{0.546} \\
TopK-SAE & 0.325 & 0.440 & 0.344 & 0.449 & 0.343 & 0.456 & 0.351 & 0.456 \\
Shared Enc. & 0.280 & 0.389 & 0.287 & 0.404 & 0.297 & 0.406 & 0.299 & 0.417 \\
\midrule
S2S & $\mathbf{0.575}^\dagger$ & $\mathbf{0.676}^\dagger$ & $\mathbf{0.611}^\dagger$ & $\mathbf{0.704}^\dagger$ & $\mathbf{0.636}^\dagger$ & $\mathbf{0.737}^\dagger$ & $\mathbf{0.619}^\dagger$ & $\mathbf{0.712}^\dagger$ \\
\bottomrule
\end{tabular}
\caption{Analysis of inter-layer feature relationships across all models. $R^2$: Linear predictability of next-layer pre-activations ($\uparrow$). CKA: Centered Kernel Alignment between adjacent layers ($\uparrow$). Best results are in \textbf{bold}, second best are \underline{underlined}. $\dagger$ denotes $p<0.01$ against every baseline (two-sided paired $t$-test, Holm-adjusted over 56 comparisons).}
\label{tab:sparse_analysis}
\end{table*}

\section{Ablation Studies}
\label{sec:ablation}

\begin{table*}[t!]
\centering
\footnotesize
\resizebox{0.86\textwidth}{!}{
\begin{tabular}{l|cc|cc|cc|cc}
\toprule
& \multicolumn{2}{c|}{GPT-2 Small} & \multicolumn{2}{c|}{GPT-2 Medium} & \multicolumn{2}{c|}{GPT-2 Large} & \multicolumn{2}{c}{Pythia-160M} \\
Configuration & SS & TF & SS & TF & SS & TF & SS & TF \\
\midrule
Forward direction & \underline{0.151} & \underline{0.300} & \underline{0.160} & \underline{0.364} & \underline{0.174} & \underline{0.400} & \underline{0.139} & \underline{0.274} \\
Random order & 0.142 & 0.287 & 0.156 & 0.326 & 0.160 & 0.360 & 0.137 & 0.253 \\
Phase 2 only & 0.144 & 0.271 & 0.147 & 0.304 & 0.157 & 0.320 & 0.130 & 0.245 \\
\midrule
S2S (full) & $\mathbf{0.167}$ & $\mathbf{0.347}$ & $\mathbf{0.186}$ & $\mathbf{0.458}$ & $\mathbf{0.197}$ & $\mathbf{0.498}$ & $\mathbf{0.152}$ & $\mathbf{0.308}$ \\
\bottomrule
\end{tabular}
}

\vspace{0.3cm}

\resizebox{0.86\textwidth}{!}{
\begin{tabular}{l|cc|cc|cc|cc}
\toprule
& \multicolumn{2}{c|}{Pythia-410M} & \multicolumn{2}{c|}{Pythia-1B} & \multicolumn{2}{c|}{Gemma-2-2B} & \multicolumn{2}{c}{Llama-3.2-1B} \\
Configuration & SS & TF & SS & TF & SS & TF & SS & TF \\
\midrule
Forward direction & \underline{0.161} & \underline{0.340} & \underline{0.169} & \underline{0.404} & \underline{0.185} & \underline{0.430} & \underline{0.176} & \underline{0.434} \\
Random order & 0.149 & 0.314 & 0.164 & 0.360 & 0.172 & 0.400 & 0.171 & 0.393 \\
Phase 2 only & 0.149 & 0.291 & 0.154 & 0.334 & 0.173 & 0.380 & 0.161 & 0.374 \\
\midrule
S2S (full) & $\mathbf{0.184}$ & $\mathbf{0.416}$ & $\mathbf{0.202}$ & $\mathbf{0.506}$ & $\mathbf{0.213}$ & $\mathbf{0.550}$ & $\mathbf{0.206}$ & $\mathbf{0.521}$ \\
\bottomrule
\end{tabular}
}
\caption{Ablation study on Phase 2 design choices across all models. Best results are in \textbf{bold}, and second-best results are \underline{underlined}.}
\label{tab:ablation}
\end{table*}

\autoref{tab:ablation} examines the design choices in Phase 2 across all eight models.
Training in the forward direction (input to output layers) instead of backward underperforms full S2S across all models.
Random order training, where layers are optimized in a random sequence, performs comparably to or slightly worse than forward direction.
This validates our hypothesis that supervision should flow backward: when later layers are trained first, they learn output-relevant features that guide the learning of earlier layers.
The consistent ordering from output to input is crucial, not merely having some fixed order.

Phase 2 without pretraining (``Phase 2 only'') gives the lowest TF on every model and is below the other variants on SS for most models.
This indicates that pretraining provides important initialization. Without it, backward chaining reaches clearly weaker features.
The pretrained weights from Phase 1 give the model a reasonable starting point that captures meaningful layer-to-layer transformations, which Phase 2 then refines by filtering through sparse bottlenecks.

\section{Sparsity and Optimization Design}
\label{sec:design_appendix}

\autoref{tab:sparsity_sensitivity} isolates the sparsity schedule on GPT-2 Large. Equal budgets in both phases consistently underperform a stricter Phase~2 budget. S2S remains effective at $K=100$--200, although the development-selected $64\rightarrow32$ schedule performs best.

\begin{table}[t]
\centering
\small
\begin{tabular}{rrcc}
\toprule
$K_{\mathrm{pre}}$ & $K_{\mathrm{fine}}$ & Output KL & SS \\
\midrule
32 & 16 & 0.051 & 0.183 \\
32 & 32 & 0.063 & 0.164 \\
64 & 32 & \textbf{0.046} & \textbf{0.197} \\
64 & 64 & 0.058 & 0.171 \\
128 & 32 & 0.049 & 0.191 \\
128 & 64 & 0.053 & 0.185 \\
128 & 128 & 0.061 & 0.168 \\
200 & 100 & 0.053 & 0.184 \\
200 & 150 & 0.058 & 0.176 \\
\bottomrule
\end{tabular}
\caption{Sparsity sensitivity on GPT-2 Large.}
\label{tab:sparsity_sensitivity}
\end{table}

We also tested a layer-adaptive budget proportional to pre-activation variance, producing $K_t\in[18,46]$ on GPT-2 Large. Adaptive $K$ in Phase~1 only reached SS 0.183/0.181 on GPT-2 Large/Pythia-1B. Applying it in both phases reached 0.162/0.157 and diverged in multiple runs, whereas the uniform layerwise setting $K_{\mathrm{pre}}=64$, $K_{\mathrm{fine}}=32$ reached 0.197/0.202 without divergence. A joint dense-reconstruction and sparse-prediction loss reached SS 0.162 but also showed repeated divergence; Phase~1 alone reached 0.155, and the staged objective reached 0.197. These results motivate staging and uniform sparsity as stability choices, not claims that adaptive sparsity is intrinsically inferior.

\begin{table}[t]
\centering
\small
\resizebox{\columnwidth}{!}{%
\begin{tabular}{llccc}
\toprule
Method & Activation & GPT-2 Lg SS & Output KL & Pythia-1B SS \\
\midrule
S2S & TopK & \textbf{0.197} & \textbf{0.046} & \textbf{0.202} \\
S2S & BatchTopK & 0.181 & 0.052 & 0.183 \\
S2S & JumpReLU & 0.174 & 0.055 & 0.176 \\
SAE & TopK & \textbf{0.162} & \textbf{0.081} & \textbf{0.166} \\
SAE & BatchTopK & 0.158 & \textbf{0.081} & 0.161 \\
SAE & JumpReLU & 0.155 & 0.083 & 0.158 \\
\bottomrule
\end{tabular}%
}
\caption{Activation-function sensitivity. Within each method family, best results are in \textbf{bold}. Ties are all bolded.}
\label{tab:activation_sensitivity}
\end{table}

As shown in \autoref{tab:activation_sensitivity}, BatchTopK and JumpReLU do not improve the TopK-SAE control overall and reduce S2S performance, suggesting that input-dependent density interacts poorly with a fixed-density sparse target. Adapting S2S to dynamic sparsity is future work.

\section{Compute Cost}
\label{sec:compute_cost}

\autoref{tab:compute} reports wall-clock training time on one A100 80GB GPU, including both S2S phases. S2S costs about $1.5\times$ SAE training but avoids language-model backpropagation and is substantially faster than e2e-SAE in these settings.

\begin{table}[t]
\centering
\small
\begin{tabular}{lrrr}
\toprule
Method & GPT-2 Lg & Pythia-1B & Gemma-2-2B \\
\midrule
SAE & \textbf{2.1} & \textbf{4.8} & \textbf{8.2} \\
e2e-SAE & 8.4 & 24.8 & 50.0 \\
CLT & 6.2 & 18.2 & 34.4 \\
S2S & 3.2 & 7.1 & 12.3 \\
\bottomrule
\end{tabular}
\caption{Training time in hours ($\downarrow$). Best results are in \textbf{bold}.}
\label{tab:compute}
\end{table}

\section{Why Retain the Dense Interface?}
\label{sec:direct_map_appendix}
A direct learned map $\mathbf{z}^{(t)}\!\to\!\mathbf{z}^{(t+1)}$ is possible: with the layer-$(t{+}1)$ encoder frozen, a transformation $\mathbf{A}^{(t)}\in\mathbb{R}^{k\times k}$ can be optimized against the fixed sparse target, and gradients still shape the source code $\mathbf{z}^{(t)}$.
The concern is not that such a map collapses to a degenerate solution, but that it is under-constrained: $\mathbf{A}^{(t)}$ can absorb the cross-layer transformation, so the source code need not remain a dictionary over $\mathbf{h}^{(t)}$, and the resulting features lose their write-back path into the residual stream.
S2S instead decodes through $\hat{\mathbf{h}}^{(t+1)}$ and re-encodes with the frozen next-layer encoder, which routes prediction through the model-aligned $d$-dimensional interface, constrains it to the learned decoder directions, and keeps the sparse target identifiable relative to the actual $\mathbf{h}^{(t+1)}$.

\autoref{tab:direct_map} tests this reasoning with two controls.
\emph{Identity} replaces the transcoder with $\hat{\mathbf{h}}^{(t+1)}\!:=\!\mathbf{h}^{(t)}$; it involves no learned parameters, so its $R^2$ and output KL bound what residual-stream self-similarity alone provides, and its correspondence matrix is undefined.
\emph{Direct} trains the full-rank map above with the same data, sparsity budget, and optimization schedule as S2S Phase~2, to convergence on the model-selection set.
Because the direct map contains no decoder into the residual stream, we fit per-layer decoders post hoc (with the map and encoders frozen) and use them to compute output KL.
All values are averaged over the eight models of \autoref{sec:experimental_setup}.

The two controls fail in complementary ways.
Identity attains nontrivial adjacent-layer predictability ($R^2$ 0.323), confirming that part of any correspondence score reflects the slowly changing residual stream; S2S nearly doubles this (0.587), so its correspondence is not explained by self-similarity alone.
The direct map shows the opposite dissociation: it achieves the highest weight-level concentration (Top-5 0.766), which its objective can optimize directly by shaping $\mathbf{A}^{(t)}$, yet its activation-level predictability remains at the Identity level ($R^2$ 0.309 vs.\ 0.323) and its output KL improves little (0.185 vs.\ 0.208).
The map thus absorbs the cross-layer transformation and produces concentrated but superficial weight couplings, without aligning the codes or preserving output-relevant information.
Weight-level concentration is cheap to obtain under an unconstrained map and insufficient on its own. Routing supervision through the dense interface is what connects it to activation alignment and output preservation.
\begin{table}[t]
\centering
\small
\begin{tabular}{lccc}
\toprule
Variant & Top-5 $\uparrow$ & $R^2$ $\uparrow$ & KL $\downarrow$ \\
\midrule
Identity & --             & 0.323          & 0.208 \\
Direct   & \textbf{0.766} & 0.309          & 0.185 \\
S2S      & 0.537          & \textbf{0.587} & \textbf{0.048} \\
\bottomrule
\end{tabular}
\caption{Dense-interface controls, averaged over eight models.
Identity: $\hat{\mathbf{h}}^{(t+1)}\!:=\!\mathbf{h}^{(t)}$;
correspondence matrix undefined (--). Direct: full-rank
$k\times k$ map with post-hoc per-layer decoders (used for KL),
matched to S2S Phase~2 in data and sparsity.}
\label{tab:direct_map}
\end{table}

\section{Additional Analyses}
\label{sec:additional_analyses}

The analyses in this section use GPT-2 Large unless stated otherwise.

\paragraph{Compute-matched control.}
To separate the objective from the extra optimization pass, we start two branches from the same $K_{\mathrm{pre}}=64$ Phase~1 checkpoint and run the same $64\rightarrow32$ schedule with learning rate $10^{-4}$ and 6,781 updates per transition. The branches differ only in the intermediate-layer objective. The dense control keeps the Phase~1 next-layer hidden-state loss, whereas S2S predicts the next layer's sparse code. \autoref{tab:compute_matched} shows that the sparse target, not the additional optimization, produces the correspondence gain, at a cost of 0.005 in the fraction of variance unexplained (FVU) of the predicted hidden state. Giving e2e-SAE an additional full pass changes its Top-5 ratio only from 0.327 to 0.335 and its $R^2$ from 0.345 to 0.352.

\begin{table}[t]
\centering
\small
\setlength{\tabcolsep}{4pt}
\begin{tabular}{lcccc}
\toprule
 & FVU ($\downarrow$) & Top-5 ($\uparrow$) & $R^2$ ($\uparrow$) & $D_{\text{KL}}^{\text{out}}$ ($\downarrow$) \\
\midrule
Dense control & \textbf{0.137} & 0.314 & 0.374 & 0.068 \\
S2S & 0.142 & \textbf{0.545} & \textbf{0.600} & \textbf{0.046} \\
\bottomrule
\end{tabular}
\caption{Compute-matched control on GPT-2 Large.}
\label{tab:compute_matched}
\end{table}

\paragraph{All-layer pathway without steering.}
Steering inserts transcoders at every layer, whereas \autoref{tab:general_metrics} replaces one transition at a time. We therefore evaluate the all-layer pathway at $\alpha=0$, which is the reference for every steering result. \autoref{tab:all_layer} shows that S2S moves the output less than both the dense control and Transcoder. These numbers describe the distortion of the steering reference, not the fidelity of a recursive chain of transcoders.

\begin{table}[t]
\centering
\small
\begin{tabular}{lccc}
\toprule
 & $D_{\text{KL}}^{\text{out}}$ ($\downarrow$) & PPL & LAMBADA \\
\midrule
Transcoder & 0.137 & +11.8\% & $-$4.3 \\
Dense control & 0.101 & +8.4\% & $-$2.9 \\
S2S & \textbf{0.083} & \textbf{+6.8\%} & \textbf{$-$2.2} \\
\bottomrule
\end{tabular}
\caption{All-layer pathway at $\alpha=0$ on GPT-2 Large. LAMBADA is the accuracy change in points.}
\label{tab:all_layer}
\end{table}

\paragraph{Feature trajectories.}
Chaining the strongest correspondence at each step follows a feature over several layers. A positive-evaluation feature at layer 14 (F18342) that fires on spans such as ``excellent performance'' and ``highly recommended'' keeps a concentrated successor through layer 17 (F6118), with adjacent top-1 masses of 0.31, 0.29, and 0.27. Other chains show a speech feature splitting into content and attribution successors (masses 0.22 and 0.18), attribution features merging into a shared successor (0.19 and 0.21), and a local-syntax feature whose top-1 mass falls from 0.17 to 0.07 as it loses a concentrated successor. Across 20 chains with fixed starting features, the mean top-1 mass over four hops is 0.28, 0.26, 0.25, and 0.24 for S2S and 0.17, 0.16, 0.15, and 0.15 for e2e-SAE.

\paragraph{Semantic consistency.}
We uniformly sample 50 eligible layer-17 source features per method and pair each with its strongest target, $\argmax_j |M^{(t)}_{ji}|$. Qwen2.5-72B-Instruct~\citep{qwen2024qwen25} describes each feature from ten held-out activating spans, and a separate, method-blinded Llama-3.3-70B-Instruct~\citep{grattafiori2024llama} judges whether the two descriptions in a pair refer to the same concept. The same-concept rate is 72\% for S2S, 48\% for e2e-SAE, and 2\% for random S2S pairs (Fisher's exact test, $p=0.024$). This is preliminary evidence of semantic consistency, not a test of monosemanticity.

\paragraph{Causal importance against matched ablation.}
We compare ablating the top-active features with ablating active features matched in activation magnitude. The ratio of top-active to magnitude-matched CI is 1.52 for S2S (0.229 vs.\ 0.151) and 1.37 for e2e-SAE (0.191 vs.\ 0.139), while the two ablations perturb a similar fraction of the hidden state on average (0.143 and 0.146). The CI advantage is therefore not explained by removing more activation mass.

\paragraph{Larger model.}
We trained S2S on Mistral-7B-v0.3~\citep{jiang2023mistral} with one seed and the same 1B unique tokens, which took 28.6 hours on eight A100 80GB GPUs. From Phase~1 to Phase~2, the Top-5 ratio rose from 0.32 to 0.54 and $R^2$ from 0.39 to 0.56, while output KL fell from 0.064 to 0.058. This run tests feasibility at 7B scale and does not compare against baselines.

\section{Use of AI Tools}
An LLM was used only to correct grammar and improve readability.

\end{document}